\documentclass[runningheads]{llncs}

\usepackage[linesnumbered,ruled,vlined]{algorithm2e}
\usepackage{algpseudocode}
\usepackage{amsmath,amssymb,amsfonts}
\usepackage{booktabs}
\usepackage{calc}
\usepackage{cite}
\usepackage{filecontents}
\usepackage{float}
\usepackage{graphicx}
\usepackage{mathtools}
\usepackage{multirow}
\usepackage{pgfplots}
\pgfplotsset{compat=1.14}
\usepackage{pgfplotstable}
\usepackage{ragged2e}
\usepackage{textcomp}
\usepackage{url}
\usepackage{xcolor}
\usepackage{graphicx,subfigure}

\begin{document}

\title{MSD-Kmeans: A Novel Algorithm for Efficient Detection of Global and Local Outliers}
\titlerunning{MSD-Kmeans}
%

\author{Yuanyuan Wei\inst{1} \and
Julian Jang-Jaccard\inst{1} \and
Fariza Sabrina\inst{2} \and
Timothy McIntosh\inst{1}}

\authorrunning{Y. Wei et al.}

\institute{Massey University, Auckland, New Zealand
\email{y.wei1@massey.ac.nz} \email{J.Jang-jaccard@massey.ac.nz}, \email{t.mcintosh@massey.ac.nz}\and
Central Queensland University, Australia 
\email{f.sabrina@cqu.edu.au}
}
\maketitle              
%
\begin{abstract}
Outlier detection is a technique in data mining that aims to detect unusual or unexpected records in the dataset.  Existing outlier detection algorithms have different pros and cons and exhibit different sensitivity to noisy data such as extreme values. In this paper, we propose a novel cluster-based outlier detection algorithm named \textit{MSD-Kmeans} that combines the statistical method of \textit{Mean and Standard Deviation} (MSD) and the machine learning clustering algorithm \textit{K-means} to detect outliers more accurately with the better control of extreme values. There are two phases in this combination method of \textit{MSD-Kmeans}: (1) applying \textit{MSD} algorithm to eliminate as many noisy data to minimize the interference on clusters, and (2) applying \textit{K-means} algorithm to obtain local optimal clusters. We evaluate our algorithm and demonstrate its effectiveness in the context of detecting possible overcharging of taxi fares, as greedy dishonest drivers may attempt to charge high fares by detouring. We compare the performance indicators of \textit{MSD-Kmeans} with those of other outlier detection algorithms, such as \textit{MSD}, \textit{K-means}, \textit{Z-score}, \textit{MIQR} and \textit{LOF}, and prove that the proposed \textit{MSD-Kmeans} algorithm achieves the highest measure of precision, accuracy and F-measure. We conclude that \textit{MSD-Kmeans} can be used for effective and efficient outlier detection on data of varying quality on IoT devices.



\end{abstract}

\section{Introduction}


Modern taxis are equipped with networked Global Positioning System (GPS) devices, a type of IoT devices, from which sufficient information on trip time, distances, fares, routes and speeds can be collected for administrative purposes or further analysis \cite{liu2014fraud,pan2013crowd,ge2011taxi,sun2013real}. Taxis play an important role in public transport provision in urban life, plugging the gaps left by buses and trains. Due to taxi fares being calculated by distance and waiting time, longer taxi trips can lead to higher fares. Factors such as traffic congestion and urban road planning could prolong taxi trips but are beyond human control. However, some greedy drivers intentionally and fraudulently take detours to push up taxi fares and pocket in more profits. Due to increased complaints of such taxi fraud received from passengers\cite{ge2011taxi,liu2014fraud}, taxi fraud detection and regulation is becoming an essential but challenging issue. One proposed method to combat it is to monitor the big data of taxi routes and driving patterns to attempt to identify suspicious outliers deviating from the average taxi routes and fares that could indicate fraud activities \cite{ge2011taxi}. Based on the frequency of occurrence, data outliers can be either random or continuous \cite{yu2017recursive}; continuous outliers of longer taxi routes and higher fares are more likely to be caused by environmental factors such as traffic congestion, while random occasional ones are more likely to be due to fraudulent activities \cite{liu2014fraud,ge2011taxi}.

Various algorithms have been proposed and developed to detect data outliers that could possibly indicate taxi fare fraud. In previous works, the main approaches include trajectories-based and parameter-based detection. Trajectories-based detection approaches analyze data of trajectories to determine whether the vehicles from starting points to destinations have detoured unnecessarily \cite{lee2008trajectory}. The full path from the starting point to the destination can be partitioned into small segments named sub-trajectories, to compare with the dataset of full paths with different sub-trajectories to detect abnormal sub-trajectories \cite{lee2008trajectory}. The analysis of GPS trajectory data can be further enhanced with social media data, when people post complaints of taxi detours \cite{pan2013crowd}. Parameter-based approaches investigate one or more  parameters such as distances, locations and speeds \textit{etc}, to detect data deviating from the average or usual norm. For example, Speed-based Fraud Detection System could detect abnormal activities by calculating and identifying the excessively fast or slow speeds using data from the GPS database \cite{liu2014fraud}. Both trajectories-based and parameter-based approaches could assist in not only identifying taxi fare fraud but also planning urban road networks \cite{zhang2011ibat,liu2014fraud}.

\textbf{Problem Statement}.
While the proposed implementations have made significant progress in identifying data outliers, some of them have not proven their scalability on larger complicated datasets. In this study, we use a combination of statistical method \textit{MSD} and machine learning method \textit{K-means}, to further improve detection accuracy on larger datasets while minimizing the impact of noisy data. To evaluate our implementation, we used the New York City (NYC) Yellow Taxi Data, a dataset of about 1.71GB collected from registered taxis in NYC in January 2016. 

\textbf{Summary of Original Contributions}

\begin{enumerate}
	\item We present the novel \textit{MSD-Kmeans} outlier detection algorithm that combines the features from the statistical method of \textit{MSD} and the machine learning method of \textit{K-means}.
	\item We apply the \textit{MSD-Kmeans} on NYC Yellow Taxi Data dataset to identify possible taxi fare fraud then demonstrates that it can efficiently detect global and local outliers.
	\item We compare the performance of \textit{MSD-Kmeans} with other outlier detection algorithms and demonstrates that \textit{MSD-Kmeans} can effectively eliminate noisy data while achieving satisfactory detection results.
\end{enumerate}

\section{Related work}\label{sec:rel}
Outlier Detection has been implemented in previous surveys and review papers using different datasets and different algorithms.  Statistical methods were developed first, to measure how each individual piece of data deviate from the statistical norm or average values; its effectiveness to detect anomaly can largely depend on the model design and means of data analysis \cite{hodge2004survey,aggarwal2017probabilistic,prasad2013statistical}. Machine Learning algorithms were later developed to assist in data analysis and became a popular technique for detecting outliers, such as the cluster-based \textit{K-means} \cite{duan2009cluster} and the density-based \textit{Local Outlier Factor} (\textit{LOF}) \cite{breunig2000lof}. Cluster-based method plays a key role in data mining, especially in data partition\cite{wang2011improved} and classification and outlier detection\cite{munz2007traffic}.

\textit{K-means} is a classic clustering algorithm used in outlier detection because of easiness of implementation \cite{jain2010data}. However, \textit{K-means} can be sensitive to noisy data when used to detect outlier \cite{abbas2008comparisons}. A few studies proposed improvements of \textit{K-means} for outlier detection. In \cite{munz2007traffic}, the \textit{Network Data Mining} (NDM) method was used to extract features from packet and flow data captured in a network before performing clustering with a distance-based \textit{K-means} algorithm. In \cite{munz2007traffic}, it was processed both classification and outlier detection simultaneously, making it suitable for scalable real-time detection, but additional work must be done to determine the optimum number of clusters.

Due to the limitations of \textit{K-means}, several studies proposed to combine \textit{K-means} with other methods for better detection outcomes. In \cite{mumtaz2010novel}, the \textit{Density Based Improved K-means Clustering} (Dbkmeans) algorithm was proposed to combine \textit{K-means} and \textit{Density Based Spatial Clustering of Applications with Noise} (DBSCAN) algorithm to get the advantage of both algorithms. Although \cite{mumtaz2010novel} could better handle clusters of circularly distributed data points and slightly overlapped clusters, the study used synthetically created data, and further work is required to conduct empirical analysis using real-world data. The outcome of this hybrid methodology showed a higher precision in outlier detection. In \cite{wang2011improved}, it was proposed to improve \textit{K-means} by applying density-based detection methods and adding the discovery and processing steps of the noise data to the original algorithm. The extra pre-processing step in \cite{wang2011improved} to exclude the interference of outliers is more time-consuming when applied to larger datasets, limiting the scalability and applicability of this algorithm. In \cite{hatamlou2012combined}, a hybrid algorithm named \textit{the Gravitational Search Algorithm and K-means} (GSA-KM)  was designed to combine \textit{GSA} and \textit{K-means} for better clustering, but it required minimum number of function evaluations to reach the optimal solution. In \cite{tsai2010triangle}, the \textit{Triangle Area-based Nearest Neighbours} (TANN) method was proposed to use \textit{K-means} to acquire centroids of each cluster, before using triangle area from each cluster centroid to get new datasets and applying \textit{K-NN} classification method to classify attacks. Although the implementation achieved higher accuracy and detection rates and the lower false negative rates, the study did not discuss whether \textit{K-means} was the optimal clustering technique for \textit{TANN}, and further work is required to examine the performance of \textit{TANN} with the datasets containing different numbers of classes.

While the previous work made practical applications of outlier detection, they have not proven their scalability on larger datasets or resistance to noisy data. In this paper, we will demonstrate that the combination of both statistical outlier detection \textit{MSD} and the machine learning of the \textit{K-means} algorithm to detect anomaly could achieve more efficient outlier detection while minimizing the interference of noisy data.

\section{Introducing New MSD-Kmeans}\label{sec:outlier}
The new\textit{MSD-Kmeans} (Algorithm~\ref{alg:msd-k}) is proposed here in this paper that combines the features of \textit{MSD} (Algorithm~\ref{alg:msd-k}, step 1) and \textit{K-means} (Algorithm~\ref{alg:msd-k}, step2). Our proposal utilities \textit{MSD} to eliminate as many global outliers (extreme values) as possible to minimize their interference on efficient clustering by \textit{K-means}. Since the number of normal data points is generally greater than the number of outliers, if the extreme value can be eliminated before clustering via \textit{K-means}, the efficiency and accuracy and local optima can be improved.

In the first phase of \textit{MSD-Kmeans}, the statistical algorithm of \textit{MSD} is used to eliminate extreme value that is defined as first stage outliers. In the second phase, utilizing the remaining normal data from the \textit{MSD} method to partition into clusters by using the \textit{K-means} algorithm is conducted. Two phases of outlier detection can be processed as follows:
\begin{enumerate}
	\item {Calculating the mean value $\mu$:
		\begin{equation}\label{eq:mean}
		\mu= \frac{1}{n}\sum_{i=1}^{n}x_i
		\end{equation}
		where $x$ ia a dataset $\{x_1,x_2,,x_3,\dots,x_i\}$; $n$ is the number of dataset of fare values from source to destination.
	}
	\item {Calculating the standard deviation value $\sigma$
		\begin{equation}\label{eq:SD}
		\sigma= \sqrt{{\frac{\sum_{i=1}^{n}(x_i-u)^2}{n}}}
		\end{equation}
	}
	\item {Figure out both normal fare value dataset $N$ and global outlier dataset $S$. The formula shows as follows
		\begin{equation}\label{eq:normal}
		N>\mu-\sigma \land N<\mu+\sigma
		\end{equation}
		\begin{equation}\label{eq:outlier}
		S>\mu+\sigma \lor S<\mu-\sigma
		\end{equation}
	}
	\item {K-means clustering based on normal dataset $N$.}
\end{enumerate}

\begin{algorithm}[h!]
	\SetAlgoLined
	\DontPrintSemicolon
	
	\KwIn{$\{x_1,x_2,x_3,\dots,x_i\}$}
	\KwOut{$o$, $n$}
	\Begin{
		step $1$: ~\footnotesize\tcc*[f]{MSD for global outliers}\linebreak
		{
			Calculate $\mu$ and $\sigma$ of $\{x_1,x_2,x_3,\dots,x_i\}$\linebreak
			\ForEach{$x_k\in\{x_1,x_2,x_3,\dots,x_i\}$}
			{
				\If{($x_k < \sigma - \mu$) OR ($x_k > \sigma + \mu$)}{Remove($x_k$)}
			}
		}
		step $2$: ~\footnotesize\tcc*[f]{K-means for local outliers}\linebreak
		{
			{$C = o\{c_1,c_2,c_3,\dots,c_j\}$ (set of cluster centroids) 
			}
			\linebreak
			\ForEach{$c_i\in C$}
			{$c_i \leftarrow e_j \in \{x_1,x_2,x_3,\dots,x_i\}$ (random selection of centroids)}
			\ForEach{$x_m\in \{x_1,x_2,x_3,\dots,x_i\}$}
			{$l(x_m) \leftarrow $AverageMinDistance$(x_m,e_n) n \in \{1 ... k\}$}
			$change \leftarrow false$
			
			\While{ $changed == false$}
			{
				\ForEach{$c_p\in C$}
				{UpdateClusters($x_p$)}
				\ForEach{$x_q\in \{x_1,x_2,x_3,\dots,x_i\}$}
				{
					$dist \leftarrow $AverageMinDistance$(e_p,e_q) q \in \{1 ... k\}$
					\If{$dist \neq l(e_p)$}
					{
						$l(e_p) \leftarrow dist$
						$changed \leftarrow true$
					}
				}
			}
			Calculate $\mu$ and $\sigma$ of $\{x_1,x_2,x_3,\dots,x_i\}$\linebreak
			\ForEach{$x_s\in \{x_1,x_2,x_3,\dots,x_i\}$}
			{
				$o$ = DistanceFromCentroid($x_s$)\linebreak
				\If{$o > \mu + 1.5*\sigma$}
				{$x_s$ is Local Outlier}
			}
			
		}
	}
	\caption{MSD-Kmeans}
	\label{alg:msd-k}
\end{algorithm}

In our implementation of \textit{MSD-Kmeans}, we used 1 standard deviation and the mean value to fence in the normal values and to fence out the global outliers $S$ (\textit{formula}~\ref{eq:outlier}). After the global outliers had been eliminated by \textit{MSD} algorithm, the remaining normal data and local outliers $N$ (\textit{formula}~\ref{eq:normal}) were grouped into two clusters by applying \textit{K-means} clustering algorithm.
 
The \textit{K-means} algorithm (Algorithm~\ref{alg:msd-k}, step2) is demonstrated here to detect local outlier values (\textit{e.g.} a value that may be within the normal range of the entire dataset but unusually low or high against surrounding values \cite{zhang2010outlier}), by assigning data into clusters with the closest mean values; those deviating from the mean value by more than 1.5 times the standard deviation are considered local outliers \cite{prasad2013statistical}. Here, we applied \textit{K-means} as second stage to detect local and the remaining global outliers among fare values. This method was implemented in Python using the $sklearn.cluster.KMeans$ module to group the dataset into $k$ clusters. If the dataset assumed that all data points in each clustering are closed to each other, outliers can be detected in each cluster based on the threshold of each cluster. The threshold in this research is calculated based on the intra-cluster distance of each cluster. Intra-cluster distance is the \textit{Euclidean Distance} calculated from each data point (fare value) to the centroid fare value of the cluster. 

According to \cite{karypis2000comparison} and our experiments, $k = 2$ appeared to have produced the best clustering results comparing to using 3 or more clusters, when some clusters could end up containing too many extreme values and affect the calculation of mean and standard deviation values. According to step $2$ in Algorithm~\ref{alg:msd-k}, the intra-cluster distance, from each data point to the centroid of the cluster it belongs to, is calculated. All intra-cluster distances are sorted into descending orders in each cluster. Finally, the threshold of local outlier values has been calculated, which is the sum of the mean value and $1.5$ times the standard deviation of intra-cluster distances in each cluster.

\section{Experiment Results}\label{sec:exp}
We evaluates our proposed \textit{MSD-Kmeans} with the NYC Yellow Taxi Data dataset, and demonstrates it was effective in identifying out-of-ordinary taxi fares which can warrant further administrative investigations. 

\subsection{Identifying Sources and Destinations}
The correct identification of pick-up (source) and drop-off (destination) locations is essential in calculating the routes traveled by taxis in order to estimate the expected fare values. Many source and destination locations have been collected by GPS devices from taxi drivers and made available in the NYC Taxi dataset. Defining outliers in those multiple sources and destinations was a challenging task, as the same source could be paired with different destinations and \textit{vice versa}. In order to detect outliers from source to destination, all fare charges collected from a pair of two blocks of the area as the source (Lower Manhattan suburb of SOHO) and destination (John F. Kennedy International (JFK) Airport) places, and obtained 79,954 records. The 79,954 records included both global and local outliers, random and continuous, and were processed using our \textit{MSD-Kmeans}. 


%

\subsection{Data Analysis}

In this experiment, we are interested in random local outliers that are more likely to be due to taxi fare fraud. The outlying fare data in this dataset could either be global outliers, possibly due to error in data collection, or local outliers that could be caused by prolonged trips. As many global outliers as possible were to be identified and removed using \textit{MSD} as part of the data cleansing before proceeding to the second stage of \textit{K-means}. The continuous outliers could be due to traffic congestion or traffic control, whereas random outliers could actually indicate an individual decision by a taxi driver to detour unnecessarily to hike fares. When using \textit{K-means} to perform clustering on the fare values, random outliers are more likely to stand out than continuous outliers.



\begin{figure}
	\centering
	\begin{minipage}[t]{.45\textwidth}
		\centering
		\includegraphics[width=1\textwidth,height=4cm]{msd}
		\caption{Normal Values and Global Outliers of Fare Value Found by MSD}
		\label{fig:msd}
	\end{minipage}\hfill
	\begin{minipage}[t]{.45\textwidth}
		\centering
		\includegraphics[width=1\textwidth,height=4cm]{msdn}
		\caption{Normal Fare Value Distribution Based on MSD Algorithm}
		\label{fig:msdn}
	\end{minipage}
\end{figure}


%
%

\begin{figure}
	\centering
	\begin{minipage}[t]{.45\textwidth}
		\centering
		\includegraphics[width=1\textwidth,height=4cm]{km1}
		\caption{Normal and Outliers Fare Value Distribution in Cluster 1 by Using K-means Algorithm}
		\label{fig:km1}
	\end{minipage}\hfill
	\begin{minipage}[t]{.45\textwidth}
		\centering
		\includegraphics[width=1\textwidth,height=4cm]{km2}
		\caption{Normal and Outliers Fare Value Distribution in Cluster 2 by using K-means Algorithm}
		\label{fig:km2}
	\end{minipage}
\end{figure}

%

\subsection{Results and Evaluation}

It was found that the majority of the trips from SOHO to JFK was between 10 to 20 miles, and most trips cost between USD \$36 to \$60 (\textit{Fig.}~\ref{fig:msd}). As can be seen in \textit{Fig.}~\ref{fig:msd}, the blue dots illustrated the normal fare value data distribution, while the red triangles illustrated the global outlier distribution found by using the \textit{MSD} statistical method as the first stage of \textit{MSD-Kmeans}. The remaining fare value data (\textit{Fig.}~\ref{fig:msdn}) were later processed with \textit{K-means} as the second stage of \textit{MSD-Kmeans}. In the second stage, \textit{K-means} clustered all data into 2 clusters. $Cluster 1$ (\textit{Fig.}~\ref{fig:km1}) collected the normal data (blue dots), some local outliers and the remaining global outliers (red triangles); the centroid fare value was calculated as \$51 and the threshold of intra-cluster distances was calculated as 8.87. $Cluster 2$ (\textit{Fig.}~\ref{fig:km2}) collected the rest of the normal fare value (blue dots) with the other local outliers (red triangles); the centroid value was calculated as \$7.7 and the threshold of intra-cluster distances was calculated as 10.83. The result of the 2 clusters was then aggregated together, and there were 5.11\% outliers in total fare amount.

\begin{figure}[hb!]\small
	\centering
	\includegraphics[width=80mm]{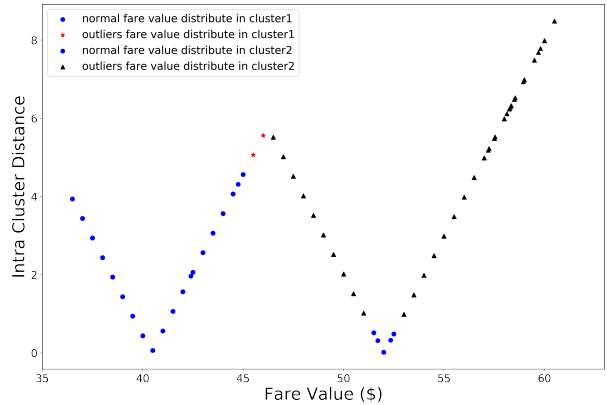}
	\caption{Normal and Outliers Fare Value Distribution by Using MSD-Kmeans Algorithm}
	\label{fig:msdk}
\end{figure}

When the results of stage 1 \textit{MSD} and stage 2 \textit{K-means} were combined together, we obtained a set of data only containing normal fare values and local fare outliers (\textit{Fig.}~\ref{fig:msdk}), which was the result of applying our \textit{MSD-Kmeans}. The new outlier threshold of intra-cluster distances was 4.58 in $cluster 1$ and 0.92 in $cluster 2$, while the centroid fare value was \$52 in $cluster 1$ and \$40 in $cluster 2$. The blue dots indicated normal fare values both in $cluster 1$ and $cluster 2$, whereas the red stars indicated outliers in $cluster 1$ and the black triangles in $cluster 2$. 

Compared with \textit{K-means} algorithm alone in \textit{Fig.}~\ref{fig:km1} and \textit{Fig.}~\ref{fig:km2}, \textit{MSD-Kmeans} obtained shorter intra-cluster distances due to the fare values closer to each other, shown in \textit{Fig.}~\ref{fig:msdk}. In other words, the further away a data point is from the centroid, it is more likely to be considered as an local outlier. \textit{K-means} clustering method identified the lowest percentage of outliers (Table~\ref{table:outlierpercentage}), because it was sensitive to the influences of global outliers but not efficient in identifying or eliminating them. In addition, choosing the optimal number $k$ of clusters is a challenging issue; often $k$ has to be determined in experiments when the value is chosen to achieve the highest clustering efficiency and the best clustering results \cite{abbas2008comparisons}. The \textit{MSD} stage of our \textit{MSD-Kmeans} algorithm is also sensitive to extreme values because the mean and standard deviation values can be shifted by the presence of those values \cite{hodge2004survey}.


The proposed method of \textit{MSD-Kmeans} is better for reducing extreme values and performing high-efficiency clustering. Using \textit{MSD-Kmeans}, there were 11.14\% outliers identified compared to \textit{MSD} (9.75\% identified) or \textit{K-means} (5.11\% identified). Despite that \textit{LOF} identified the highest number of outliers (19.72\% identified), the precision (26.1\%) and accuracy (38.0\%) were lower than those of other algorithms (Table~\ref{table:PerformanceComparison}).

\begin{table}[!h]
	\centering
	\caption{Implementation Results of Different Algorithms}
	\label{table:outlierpercentage}
	\begin{tabular}{|c|c|c|c|c|c|c|}
		\hline
		\textbf{Algorithm} & \textbf{\begin{tabular}[c]{@{}c@{}}Total\\ Number\\of Records\end{tabular}} & \textbf{\begin{tabular}[c]{@{}c@{}}Normal\\Records\\Found\end{tabular}} & \textbf{\begin{tabular}[c]{@{}c@{}}Outliers\\Found\end{tabular}} & \textbf{Outliers (\%)}  \\ \hline
		MSD & 79,954 & 71,416 & 7,799 & 9.75 \\ \hline
		Z-score & 79,954 & 75,460 & 4,494 & 5.62  \\ \hline
		MIQR & 79,954 & 69,418 & 10,473 & 13.10  \\ \hline
		K-means & 79,954 & 75,864 & 4,090 &  5.11 \\ \hline
		LOF & 261 & 218 & 43 & 19.72 \\ \hline
		MSD-Kmeans &79,954 & 71,044 & 8,910 & 11.14 \\ \hline
		
	\end{tabular}
\end{table}

\noindent
\begin{table*}[!h]
	\centering
	\caption{Performance Comparison of Outlier Detection Algorithms using NYC Taxi Dataset}
	\label{table:PerformanceComparison}
	\begin{tabular}{|c|c|c|c|c|c|c|c|c|}
		\hline
		\textbf{\begin{tabular}[c]{@{}c@{}}Outlier \\Detection \\Algorithm\end{tabular}} & \textbf{\begin{tabular}[c]{@{}c@{}}TPR \\(\%)\end{tabular}} & \textbf{\begin{tabular}[c]{@{}c@{}}FPR \\(\%)\end{tabular}} & \textbf{\begin{tabular}[c]{@{}c@{}}Precision \\(\%)\end{tabular}} & \textbf{\begin{tabular}[c]{@{}c@{}}Accuracy \\(\%)\end{tabular}} & \textbf{\begin{tabular}[c]{@{}c@{}}Recall \\(\%)\end{tabular}} & \textbf{\begin{tabular}[c]{@{}c@{}}F-measure \\(\%)\end{tabular}} & \textbf{\begin{tabular}[c]{@{}c@{}}Execution \\Time (MS)\end{tabular}} \\ \hline
		MSD & 99.9 & 24.2 & 96.6 & 96.9 & 99.9 & 98.2 & 21\\ \hline
		Z-score & 100 & 48.9 & 94.3 & 94.6 & 100 & 97.1 & 157\\ \hline
		MIQR & 97.8 & 12.6 & 98.1 & 96.4 & 97.8 & 98.0 & 54 \\ \hline
		K-means\cite{duan2009cluster} & 99.7 & 55.6 & 93.5 & 93.7 & 99.7 & 96.6 & 1,132\\ \hline
		LOF\cite{breunig2000lof} &98.2 & 79.3 & 26.1 & 38.0 & 98.2 & 41.3 & 31,483\\ \hline
		MSD-Kmeans & 98.5 & 11.6 & 98.6 & 97.4 & 98.5 & 98.6 & 824\\ \hline
	\end{tabular}
\end{table*}

\section{Discussion}
In our evaluation, we demonstrated that our novel \textit{MSD-Kmeans} is a promising clustering algorithm that was efficient, accurate and resistant to the interference of extreme values. 

\subsection{Performance Comparison}
The performance of outlier detection algorithms can be evaluated using six possible performance indicators: TPR, FPR, Precision, Accuracy, Recall and F-measure \cite{fawcett2006introduction}. We have calculated our results and compared with those of other outlier detection algorithms applied on the same NYC taxi fare value data. The results are shown in Table~\ref{table:PerformanceComparison}. It was found that the \textit{MSD-Kmeans} algorithm had the lowest FPR (11.6\%; lower is better), the highest Precision (98.6\%), Accuracy (97.4\%) and F-measure (98.6\%; higher is better), although its TPR was not as high as that of \textit{MSD}, \textit{Z-Score} or \textit{K-means}, only higher than the TPR of \textit{MIQR}. \textit{MSD-Kmeans} was found to have higher Precision and Accuracy than those of \textit{MSD} or \textit{K-means} , as \textit{MSD} or \textit{K-means} alone is more sensitive to noisy data and could produce skewed results. Because the \textit{MSD} algorithm looks for outlier based on standard deviation, a large number of extreme values can increase its standard deviation values, decreasing the accuracy of \textit{MSD}. \textit{K-means} algorithm itself suffers the similar issue of sensitivity to noisy data. As in \textit{Fig.}~\ref{fig:km2}, the fare amount is from $0$ to $30$ dollars. However, based on criteria fare amount from $42$ to $62$ dollars, the whole cluster 2 is defined as outlier. It demonstrated that too much noisy data has impact on clustering. The proposed \textit{MSD-Kmeans} combination algorithm in general performed well. As can be seen in Table~\ref{table:PerformanceComparison}, as a result, the proposed method of \textit{MSD-Kmeans} achieved the highest precision and accuracy, which means \textit{MSD-Kmeans} obtained the results more correctly.

\subsection{Improving MSD-Kmeans by Parallelizing of K-means}
The efficiency of our proposed \textit{MSD-Kmeans} can be further improved by parallelizing \textit{K-means}, since \textit{MSD-Kmeans} is performed in two stages (\textit{MSD} and \textit{K-means} respectively). Parallel computing is a technique to divide a larger problem into smaller problems, to carry out the execution of computation simultaneously on more than one computation unit, and to aggregate the final results back to one in the end \cite{kumar1994introduction}. Parallel programming has been introduced to data mining and processing, and many algorithms have adopted to be parallelized \cite{chen2015data}. To our best knowledge, parallel implementations of \textit{MSD} have not been found in current literature, which could be an ongoing challenge for future researchers. However, parallelizing \textit{K-means} has been well-researched in \cite{farivar2008parallel,kantabutra2000parallel,zhang2006study,zhao2009parallel}, achieving from twice the efficiency \cite{zhang2006study} on CPU to 68 times on GPU \cite{farivar2008parallel} when the number of clusters $k$ is 2. As can be seen in table~\ref{table:PerformanceComparison}, the parallel performance run-time in \textit{MSD-Kmeans} algorithm cost less time consumption ($842ms$), compared to other machine learning algorithms here, such as \textit{K-means} ($1,132ms$) and \textit{LOF} ($31,483ms$). Many modern IoT devices are now equipped with multi-core CPUs and GPUs. It may be possible to boost the efficiency of \textit{MSD-Kmeans} by parallelizing the second stage of \textit{K-means} computation. This is outside the scope of this paper and would require further research and testing.

In summary, \textit{MSD-Kmeans} can obtain better results in detecting outliers than \textit{MSD} alone or \textit{K-means} alone. However, in this research, \textit{MSD-Kmeans} is used for detecting outliers based on univariate of fare value. The challenge to detect multivariate of other features can be attempted in further work.

\section{Conclusion}\label{sec:con}
In this paper, we proposed a novel outlier detection algorithm named \textit{MSD-Kmeans} that combined \textit{MSD} and \textit{K-means} to detect global and local outlier values, and applied it to identify outlying amounts of taxi fares based on distance traveled in NYC taxi dataset, which could indicate taxi fare fraud. We also applied a few other algorithms including \textit{MSD} and \textit{K-means} algorithms individually to the same dataset to compare the results. The \textit{MSD} algorithm as a statistical method can be used in not only outlier detection but also identifying extreme values, but is sensitive to the presence of extreme values. The \textit{K-means} algorithm is a machine learning based algorithm but suffers similar issues. Our proposed hybrid method \textit{MSD-Kmeans} applies the \textit{MSD} algorithm to eliminate as many extreme values as possible, before applying \textit{K-means} clustering algorithm to cluster normalized dataset in different groups. Our experimental result demonstrated that \textit{MSD-Kmeans} achieved the best precision, accuracy, and F-measure, with the lowest false positive rate, compared to other outlier detection algorithms applied on the same dataset. We believe \textit{MSD-Kmeans} is a promising algorithm in outlier detection that could benefit processing of sensor data\cite{weyers2017low} from networked IoT devices. Further work could be done to different datasets to test its practicability and scalability.

\bibliographystyle{splncs04}
\bibliography{bibfile}

\begin{thebibliography}{10}
\providecommand{\url}[1]{\texttt{#1}}
\providecommand{\urlprefix}{URL }
\providecommand{\doi}[1]{https://doi.org/#1}

\bibitem{abbas2008comparisons}
Abbas, O.A.: Comparisons between data clustering algorithms. International Arab
  Journal of Information Technology (IAJIT)  \textbf{5}(3) (2008)

\bibitem{aggarwal2017probabilistic}
Aggarwal, C.C.: Probabilistic and statistical models for outlier detection. In:
  Outlier Analysis, pp. 35--64. Springer (2017)

\bibitem{breunig2000lof}
Breunig, M.M., Kriegel, H.P., Ng, R.T., Sander, J.: Lof: identifying
  density-based local outliers. In: ACM sigmod record. vol.~29, pp. 93--104.
  ACM (2000)

\bibitem{chen2015data}
Chen, F., Deng, P., Wan, J., Zhang, D., Vasilakos, A.V., Rong, X.: Data mining
  for the internet of things: literature review and challenges. International
  Journal of Distributed Sensor Networks  \textbf{11}(8),  431047 (2015)

\bibitem{duan2009cluster}
Duan, L., Xu, L., Liu, Y., Lee, J.: Cluster-based outlier detection. Annals of
  Operations Research  \textbf{168}(1),  151--168 (2009)

\bibitem{farivar2008parallel}
Farivar, R., Rebolledo, D., Chan, E., Campbell, R.H.: A parallel implementation
  of k-means clustering on gpus. In: Pdpta. vol.~13, pp. 212--312 (2008)

\bibitem{fawcett2006introduction}
Fawcett, T.: An introduction to roc analysis. Pattern recognition letters
  \textbf{27}(8),  861--874 (2006)

\bibitem{ge2011taxi}
Ge, Y., Xiong, H., Liu, C., Zhou, Z.H.: A taxi driving fraud detection system.
  In: Data Mining (ICDM), 2011 IEEE 11th International Conference on. pp.
  181--190. IEEE (2011)

\bibitem{hatamlou2012combined}
Hatamlou, A., Abdullah, S., Nezamabadi-Pour, H.: A combined approach for
  clustering based on k-means and gravitational search algorithms. Swarm and
  Evolutionary Computation  \textbf{6},  47--52 (2012)

\bibitem{hodge2004survey}
Hodge, V., Austin, J.: A survey of outlier detection methodologies. Artificial
  intelligence review  \textbf{22}(2),  85--126 (2004)

\bibitem{jain2010data}
Jain, A.K.: Data clustering: 50 years beyond k-means. Pattern recognition
  letters  \textbf{31}(8),  651--666 (2010)

\bibitem{kantabutra2000parallel}
Kantabutra, S., Couch, A.L.: Parallel k-means clustering algorithm on nows.
  NECTEC Technical journal  \textbf{1}(6),  243--247 (2000)

\bibitem{karypis2000comparison}
Karypis, M.S.G., Kumar, V., Steinbach, M.: A comparison of document clustering
  techniques. In: TextMining Workshop at KDD2000 (May 2000) (2000)

\bibitem{kumar1994introduction}
Kumar, V.: Introduction to Parallel Computing: Design and Analysing of
  Algorithms. Springer (1994)

\bibitem{lee2008trajectory}
Lee, J.G., Han, J., Li, X.: Trajectory outlier detection: A
  partition-and-detect framework. In: 2008 IEEE 24th International Conference
  on Data Engineering. pp. 140--149. IEEE (2008)

\bibitem{liu2014fraud}
Liu, S., Ni, L.M., Krishnan, R.: Fraud detection from taxis' driving behaviors.
  IEEE Transactions on Vehicular Technology  \textbf{63}(1),  464--472 (2014)

\bibitem{mumtaz2010novel}
Mumtaz, K., Duraiswamy, K.: A novel density based improved k-means clustering
  algorithm--dbkmeans. International Journal on computer science and
  Engineering  \textbf{2}(2),  213--218 (2010)

\bibitem{munz2007traffic}
M{\"u}nz, G., Li, S., Carle, G.: Traffic anomaly detection using k-means
  clustering. In: GI/ITG Workshop MMBnet. pp. 13--14 (2007)

\bibitem{pan2013crowd}
Pan, B., Zheng, Y., Wilkie, D., Shahabi, C.: Crowd sensing of traffic anomalies
  based on human mobility and social media. In: Proceedings of the 21st ACM
  SIGSPATIAL International Conference on Advances in Geographic Information
  Systems. pp. 344--353. ACM (2013)

\bibitem{prasad2013statistical}
Prasad, Y.S., Krishna, G.R.: Statistical anomaly detection technique for real
  time datasets. International Journal of Computer Trends and Technology
  (IJCTT)  \textbf{6}(2),  89--94 (2013)

\bibitem{sun2013real}
Sun, L., Zhang, D., Chen, C., Castro, P.S., Li, S., Wang, Z.: Real time
  anomalous trajectory detection and analysis. Mobile Networks and Applications
   \textbf{18}(3),  341--356 (2013)

\bibitem{tsai2010triangle}
Tsai, C.F., Lin, C.Y.: A triangle area based nearest neighbors approach to
  intrusion detection. Pattern recognition  \textbf{43}(1),  222--229 (2010)

\bibitem{wang2011improved}
Wang, J., Su, X.: An improved k-means clustering algorithm. In: Communication
  Software and Networks (ICCSN), 2011 IEEE 3rd International Conference on. pp.
  44--46. IEEE (2011)

\bibitem{weyers2017low}
Weyers, R., Jang-Jaccard, J., Moses, A., Wang, Y., Boulic, M., Chitty, C.,
  Phipps, R., Cunningham, C.: Low-cost indoor air quality (iaq) platform for
  healthier classrooms in new zealand: Engineering issues. In: 2017 4th
  Asia-Pacific World Congress on Computer Science and Engineering (APWC on
  CSE). pp. 208--215. IEEE (2017)

\bibitem{yu2017recursive}
Yu, T., Wang, X., Shami, A.: Recursive principal component analysis-based data
  outlier detection and sensor data aggregation in iot systems. IEEE Internet
  of Things Journal  \textbf{4}(6),  2207--2216 (2017)

\bibitem{zhang2011ibat}
Zhang, D., Li, N., Zhou, Z.H., Chen, C., Sun, L., Li, S.: ibat: detecting
  anomalous taxi trajectories from gps traces. In: Proceedings of the 13th
  international conference on Ubiquitous computing. pp. 99--108. ACM (2011)

\bibitem{zhang2010outlier}
Zhang, Y., Meratnia, N., Havinga, P.J.: Outlier detection techniques for
  wireless sensor networks: A survey. IEEE Communications Surveys and Tutorials
   \textbf{12}(2),  159--170 (2010)

\bibitem{zhang2006study}
Zhang, Y., Xiong, Z., Mao, J., Ou, L.: The study of parallel k-means algorithm.
  In: 2006 6th World Congress on Intelligent Control and Automation. vol.~2,
  pp. 5868--5871. IEEE (2006)

\bibitem{zhao2009parallel}
Zhao, W., Ma, H., He, Q.: Parallel k-means clustering based on mapreduce. In:
  IEEE International Conference on Cloud Computing. pp. 674--679. Springer
  (2009)

\end{thebibliography}

\end{document}